\documentclass[conference, 9pt]{IEEEtran}
\usepackage[letterpaper, left=1in, right=1.1in, top=0.75in, bottom=1.1in]{geometry}
\usepackage{cite}
\usepackage{amsmath,amssymb,amsfonts,nccmath}
\usepackage{adjustbox}
\usepackage{multirow}
\usepackage[table,xcdraw]{xcolor}
\usepackage{xcolor}
\usepackage{url}
\usepackage{footnote}
\usepackage{float}
\usepackage{comment}

\DeclareRobustCommand*{\IEEEauthorrefmark}[1]{%
  \raisebox{0pt}[0pt][0pt]{\textsuperscript{\footnotesize #1}}%
}

\def\BibTeX{{\rm B\kern-.05em{\sc i\kern-.025em b}\kern-.08em
    T\kern-.1667em\lower.7ex\hbox{E}\kern-.125emX}}
    
\begin{document}
\title{IBB Traffic Graph Data: Benchmarking and Road Traffic Prediction Model}

\author{%
  \IEEEauthorblockN{%
    Eren Olug\IEEEauthorrefmark{1,3,4},
    Kiymet Kaya\IEEEauthorrefmark{2,3,4}, 
    Resul Tugay\IEEEauthorrefmark{5},
    and
    Sule Gunduz Oguducu\IEEEauthorrefmark{1,3}%
  }%
  \IEEEauthorblockA{\IEEEauthorrefmark{1} Istanbul Technical University, Department of Artificial Intelligence and Data Engineering, Istanbul, Turkey}
  \IEEEauthorblockA{\IEEEauthorrefmark{2} Istanbul Technical University, Department of Computer Engineering, Istanbul, Turkey}
  \IEEEauthorblockA{\IEEEauthorrefmark{3} ITU AI Research and Application Center, Istanbul, Turkey}
  \IEEEauthorblockA{\IEEEauthorrefmark{4} BTS Group, Istanbul, Turkey}
\IEEEauthorblockA{\IEEEauthorrefmark{5} Gazi University, Department of Computer Engineering, Ankara, Turkey} 
Email: olug20@itu.edu.tr, kayak16@itu.edu.tr, resultugay@gazi.edu.tr, sgunduz@itu.edu.tr
}
\maketitle
\begin{abstract}

Road traffic congestion prediction is a crucial component of intelligent transportation systems, since it enables proactive traffic management, enhances suburban experience, reduces environmental impact, and improves overall safety and efficiency. Although there are several public datasets, especially for metropolitan areas, these datasets may not be applicable to practical scenarios due to insufficiency in the scale of data (i.e. number of sensors and road links) and several external factors like different characteristics of the target area such as urban, highways and the data collection location. To address this, this paper introduces a novel IBB Traffic graph dataset as an alternative benchmark dataset to mitigate these limitations and enrich the literature with new geographical characteristics. IBB Traffic graph dataset covers the sensor data collected at 2451 distinct locations. Moreover, we propose a novel Road Traffic Prediction Model that strengthens temporal links through feature engineering, node embedding with GLEE to represent inter-related relationships within the traffic network, and traffic prediction with ExtraTrees. The results indicate that the proposed model consistently outperforms the baseline models, demonstrating an average accuracy improvement of 4\%.
\end{abstract}


\section{Introduction} \label{sec:intro}
The importance of intelligent transportation systems (ITSs) has rapidly increased as they enhance travel efficiency, reduce traffic incidents, and improve safety. 
Traffic prediction involves predicting future traffic behavior in a given area according to current and past traffic conditions. This information is crucial for multiple purposes, such as reducing congestion, optimizing transportation systems, and enhancing road safety. 

With the growing capabilities of road sensor devices, various information about traffic networks, such as flow and speed, can be collected and prepared for several downstream tasks not only for traffic predicting, including forecasting of traffic flow, travel time, and traffic density, but also other real-world applications such as travel demand prediction for ride-hailing services, autonomous vehicles, intersection management, parking management, urban planning, and transportation safety \cite{10077454}. Traditionally, traffic prediction has utilized methods like rule-based models and time-series analysis. However, these approaches often fall short of capturing the full complexity, variability of traffic patterns in graph data.

Traffic flow prediction for a specific road is a difficult problem because it is influenced by many factors, including the day, time of day, weather conditions, and the traffic flow of neighboring roads. Thus, traffic prediction requires generating data from the non-Euclidean domain, necessitating the use of graph data structures to represent complex spatial relationships and interdependencies between objects \cite{9046288}. This task is particularly challenging due to the spatiotemporal nature of traffic network data, which encapsulates both intricate spatial relationships and dynamic temporal dependencies \cite{jin2023survey, yin}. 

Graph Neural Networks (GNNs) can directly process the graph structure (i.e., adjacency information) along with node (sensors in roads) features. Numerous techniques for extracting features from graphs \cite{Torres_2020,narayanan2017graph2vec, Perozzi_2014, Donnat_2018} have been shown to be effective in handling graph structures, including their adjacency information. These methods enable the extraction of node embeddings of the graph in Euclidean space.

Publicly available road traffic prediction datasets differ in several ways (1) size: the number of sensors or data points, (2) temporal coverage: how long period the data covers, (3) sensor metadata: information about the sensors (e.g. location, type), and (4) road type: whether the data is collected on highways or urban streets. One major limitation in many popular datasets, like METR-LA\cite{li2018diffusionconvolutionalrecurrentneural}, is their focus on only highways. This focus results in the lack of ability to recognize urban traffic networks successfully. Another limitation is that these datasets are often scale-limited compared to real-world road networks, making it harder to train complex forecasting models. For instance, METR-LA and PeMS-BAY\cite{liu2023largest}, two of the most widely used datasets, contain only 207 and 325 sensors, respectively. Additionally, only a few of the datasets \cite{liu2023largest} cover a long enough temporal period, which is crucial in capturing long-term traffic patterns. Considering the impact of data quality on model performance, these limitations reduce the effectiveness of traffic prediction models.

In this study, we aim to enrich the literature by introducing the IBB Traffic graph dataset, which covers 2,451 sensors located in Istanbul, Turkey, including both urban and highway roads. Given the features collected by sensors, this dataset can be utilized not only for predicting road traffic anomalies but also for forecasting speed, flow, and travel time. Moreover, we propose a novel Road Traffic Prediction Model that strengthens temporal links through feature engineering, node embedding with GLEE to represent inter-related relationships within the traffic network, and traffic prediction with ExtraTrees. 

The main contributions of our study can be summarized as follows:
\begin{itemize}
    \item \textbf{A novel IBB traffic graph dataset} comprises sensor traffic data collected from 2,451 distinct locations over a four-year period with measurements taken at one-hour intervals.    
    \item  \textbf{A novel Road Traffic Prediction Model} that strengthens temporal links through feature engineering, node embedding with GLEE to represent inter-related relationships within the traffic network, and traffic prediction with ExtraTrees.

\end{itemize}

The rest of the paper is organized as follows. Section \ref{sec:2} presents related works on traffic congestion prediction problem. Section \ref{sec:3} gives details of the IBB traffic graph data generation process. Section \ref{sec:4} gives details of the proposed methodology. Section \ref{sec:5} presents the experimental results of the study. Lastly, Section \ref{sec:6} concludes the paper.

\section{Literature Review} \label{sec:2} 

Extreme Gradient Boosting (XGBoost) is proposed for higher resolution traffic state estimation using the origin-destination relationship of segment flow data between upstream and downstream on the highway \cite{sun2021spatio}. The performance of machine learning methods such as linear regression, Random Forest (RF), and ARIMA is evaluated for traffic congestion prediction, which is one of the vital components of Intelligent Transportation Systems in smart cities \cite{bai2021prepct}. The growth of big data makes it possible to study traffic congestion issues through data from sensors in various locations. Considering rainfall data as a traffic congestion indicator, a stacked generalization model suitable for big and complex data using RF, CatBoost, and XGBoost algorithms is proposed \cite{tran2022predicting}. LightGBM model proposed for Freeway Short-Term Travel Time Prediction is proved to be more accurate and has a shorter model training time than KNN and Gradient Boosting Machine (GBM) \cite{wang2021freeway}.

Models trained and evaluated on widely used traffic datasets may not be applicable to practical scenarios due to the limited scale of these datasets compared to real-world traffic networks \cite{liu2023largest}. In other words, widely used datasets such as PeMS03, PeMS04, PeMS07, PeMS08, METR-LA, and PeMS-BAY\cite{Song_Lin_Guo_Wan_2020, li2018diffusionconvolutionalrecurrentneural} contain a low number of sensors, resulting in a small road network with an insufficient number of nodes. In this study, we enrich the literature by introducing the IBB Traffic graph dataset, which covers 2451 sensors located in Istanbul, including both urban and highway roads. 

\section{IBB Graph-Based Road Network Dataset} \label{sec:3} 

Traffic in Istanbul is notable due to the city's unique characteristics, including its distinctive geographical features and its status as a metropolitan area with over five million vehicles\cite{tuik}. The city spans two large landmasses, each located on a different continent (Asia and Europe), separated by the Bosporus Strait. In addition to intra-continental traffic flow, millions of people travel between continents daily using three different bridges connecting these continents.

\begin{figure} [htbp]
    \centering
    \includegraphics[width=.99\linewidth]{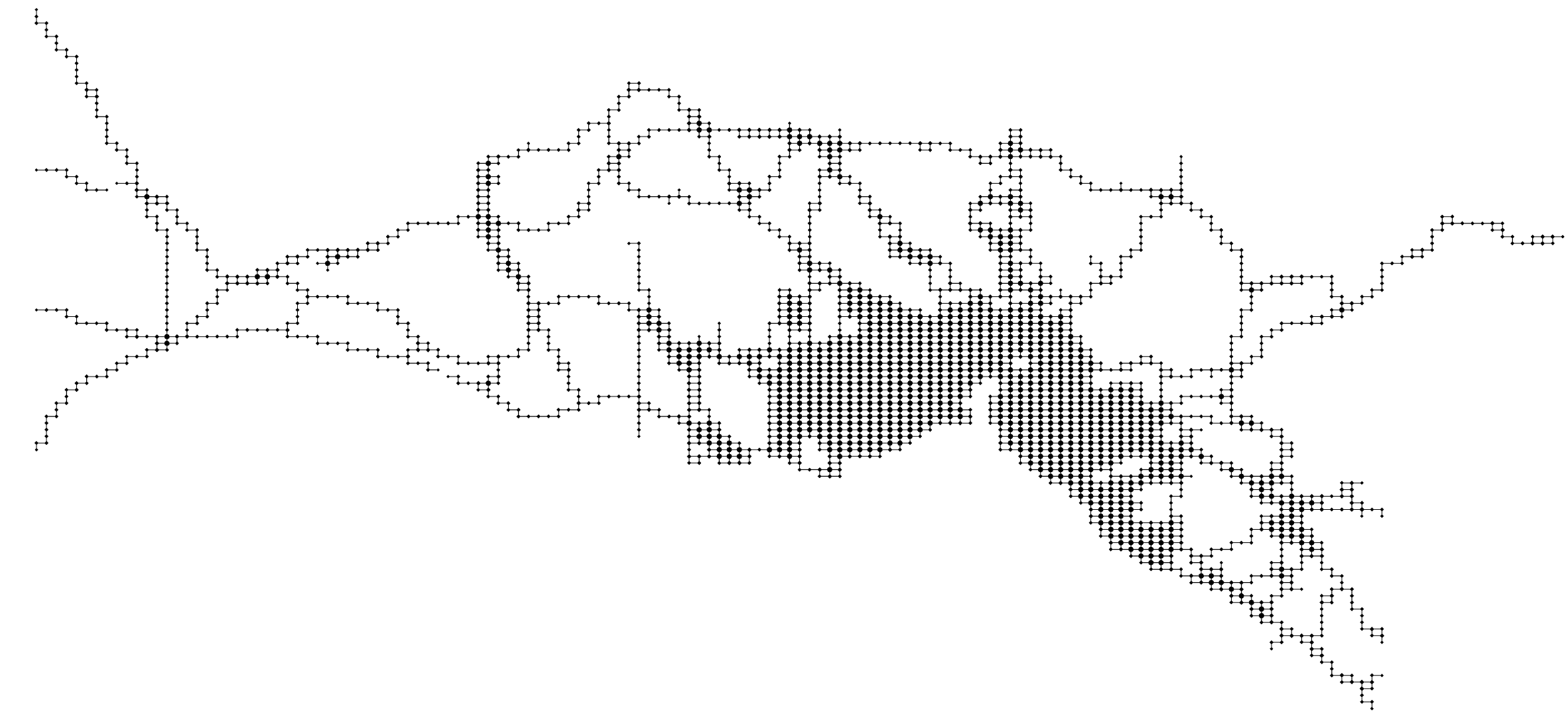}
    \caption{Istanbul Road Traffic Network}
    \label{fig:geolayout}
\end{figure}

The proposed IBB Traffic graph dataset encompasses four years of traffic data, including features such as traffic state (e.g., flow and speed) and sensor information. The traffic data-collecting sensors are strategically positioned at approximately equal intervals throughout the city, covering both highways and urban areas. In the IBB Traffic graph dataset, sensor measurement frequency is one hour, yielding 24 measurements per day for each of the 2451 sensors. Since edge connections are not explicitly defined, we connected two nodes with an unweighted edge if the corresponding sensors were adjacent to each other. This approach results in a total of 6667 edges and an average degree of approximately 2.7 in the road network. The topology of the resulting data is visualized in geolayout representation in Figure \ref{fig:geolayout} to provide an insight into the IBB Traffic network.

\begin{figure*}[htbp]
\includegraphics[width=.99\linewidth]{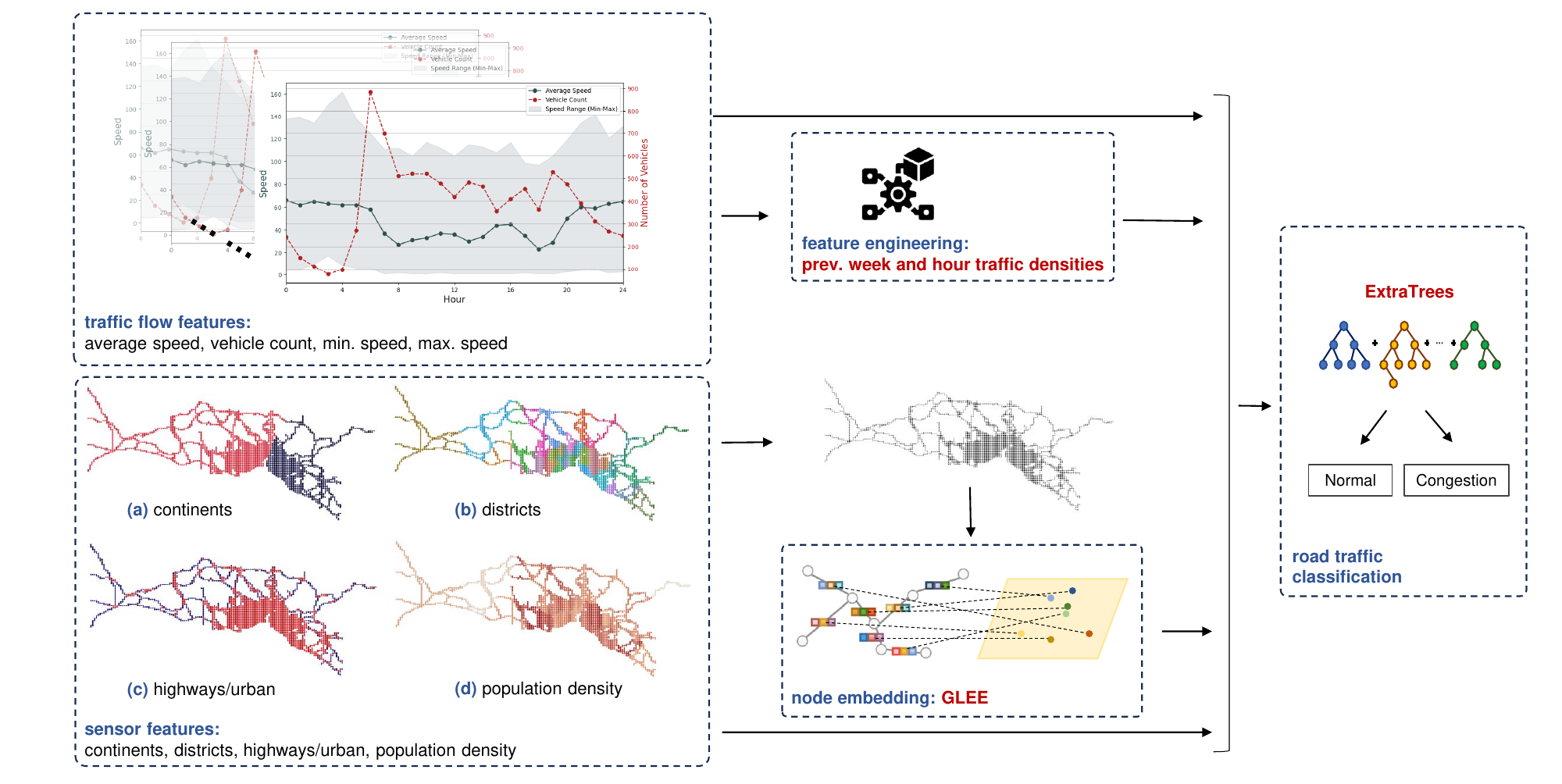}
\caption{Road Traffic Prediction Model}
\label{fig_proposed}
\end{figure*}

In the graph representation of the IBB Traffic dataset, all features are encoded as node features meaning the edges are non-attributed. The node features consist of two categories: traffic flow features and sensor features. As the name implies, traffic features are collected from traffic flow measured by sensors in that time range such as \textit{minimum speed}, \textit{maximum speed}, \textit{average speed}, and \textit{number of passing vehicles}. On the other hand, sensor features include spatial information about sensors such as \textit{continent}, \textit{district}, \textit{whether the sensor is on a highway or not}, and \textit{the population density} around the sensor.

\section{Methodology} \label{sec:4} 

The proposed Road Traffic Prediction Model is illustrated in Figure \ref{fig_proposed}. The model benefits of traffic flow features: average speed, vehicle count, minimum speed, and maximum speed, and sensor features encompass geographical categorizations like continents, districts, highways/urban areas, and population density, which are shown in maps. With feature engineering study, previous week and hour traffic densities are gathered. Moreover, node embeddings are obtained with the GLEE method. All of the features are then processed and combined by the ExtraTrees classifier to distinguish between normal and congestion traffic. Next, we give details of the node embedding method GLEE and traffic congestion classification method ExtraTrees. The implementation details and the code repository of the proposed model are available here\footnote{https://github.com/erenolg/IBBTraffic}.

\subsection{Node Embedding: GLEE} 

GLEE \cite{Torres_2020} is closely related to the Laplacian Eigenmaps (LE) method. LE constructs node embeddings based on the spectral properties of the Laplacian matrix of the graph. However, while LE aims to minimize the distance between similar nodes based on the spectral properties of the Laplacian matrix, GLEE disposes of the distance-minimization assumption and uses the Laplacian matrix to find embeddings with geometric properties, by leveraging the simplex geometry of the graph.

Let $\Lambda$ be the diagonal matrix having eigenvalues of $\textbf{L}$ as entries, GLEE solves a distance maximization problem given in Equation \ref{glee_e0}, where the objective function is $\operatorname{tr}(\textbf{Y}^T \textbf{L} \textbf{Y})$ under the constraint $\textbf{Y}^T \textbf{Y} = \Lambda$.

\begin{equation}
\arg \max_{\textbf{Y} \in \mathbb{R}^{n \times d}} \operatorname{tr}(\textbf{Y}^T \textbf{L} \textbf{Y}) \quad \text{s.t.} \quad \textbf{Y}^T \textbf{Y} = \Lambda
\label{glee_e0}
\end{equation}

More concretely, given a set of data points \(\{x_1, x_2, \ldots, x_n\}\), the first step in Geometric Laplacian Eigenmap Embedding (GLEE) is to construct a graph \(G = (V, E)\) where each node represents a data point. The edge weight matrix $\textbf{A}$, is typically computed using a Gaussian kernel as in Equation \ref{glee_1}. Next, the degree matrix $\textbf{D}$ is computed, which is a diagonal matrix represents the sum of the weights of the edges connected to node \(i\).

\begin{equation}
A_{ij} = \exp\left(-\frac{\|x_i - x_j\|^2}{2\sigma^2}\right).
\label{glee_1}
\end{equation}

Given the weight matrix \textbf{A} and the degree matrix \textbf{D}, we then compute the $\textbf{L}$, defined as $\textbf{L} = \textbf{D} - \textbf{A}$. The core of GLEE involves solving the generalized eigenvalue problem $\textbf{L} \mathbf{y} = \lambda \textbf{D} \mathbf{y}$, where \(\mathbf{y}\) are the eigenvectors and \(\lambda\) are the eigenvalues. The solution to this eigenvalue problem provides a set of eigenvectors \(\mathbf{y}_1, \mathbf{y}_2, \ldots, \mathbf{y}_m\) corresponding to the smallest eigenvalues. Finally, the data points \(x_i\) are embedded into a lower-dimensional space by using the eigenvectors as coordinates $y_i = (y_i^{(1)}, y_i^{(2)}, \ldots, y_i^{(m)})$, where \(y_i^{(j)}\) is the \(i\)-th component of the \(j\)-th eigenvector. This process results in a low-dimensional representation of the original high-dimensional data, preserving the geometric structure of the data as captured by the graph.

\begin{table*}[!htbp]
\caption{Traffic Congestion Classification Results}
\renewcommand{\arraystretch}{1.2}
\label{tab:res1}
\centering
\resizebox{.9\linewidth}{!}{
\begin{tabular}{lccccclccccc}
\hline
\multicolumn{1}{c}{} & \multicolumn{5}{c}{\textbf{wo/ feature enginnering}} &  & \multicolumn{5}{c}{\textbf{w/ feature enginnering}} \\ \cline{2-12} 
\multicolumn{1}{c}{\multirow{-2}{*}{\textbf{\begin{tabular}[c]{@{}c@{}}Classification\\ Model\end{tabular}}}} & \textbf{Acc.} & \textbf{Precision} & \textbf{Recall} & \textbf{F1} & \textbf{AUROC} & \multicolumn{1}{c}{\textbf{}} & \textbf{Acc} & \textbf{Precision} & \textbf{Recall} & \textbf{F1} & \textbf{AUROC} \\ \hline
\textbf{LR} & 0.7739 & \cellcolor[HTML]{C3E2C3}\textbf{0.6250} & 0.2066 & 0.3105 & 0.8005 &  & \cellcolor[HTML]{C3E2C3}\textbf{0.9572} & 0.9310 & \cellcolor[HTML]{C3E2C3}\textbf{0.8925} & \cellcolor[HTML]{C3E2C3}\textbf{0.9113} & \cellcolor[HTML]{C3E2C3}\textbf{0.9920} \\ \hline
\textbf{KNN} & 0.7596 & 0.5180 & \cellcolor[HTML]{C3E2C3}\textbf{0.3553} & 0.4215 & 0.7491 &  & 0.9490 & 0.9000 & \cellcolor[HTML]{C3E2C3}\textbf{0.8925} & 0.8962 & 0.9625 \\ \hline
\textbf{RF} & 0.7820 & 0.5972 & \cellcolor[HTML]{C3E2C3}\textbf{0.3553} & \cellcolor[HTML]{C3E2C3}\textbf{0.4456} & 0.8396 &  & 0.9551 & 0.9230 & \cellcolor[HTML]{C3E2C3}\textbf{0.8925} & 0.9075 & 0.9829 \\ \hline
\textbf{XGBoost} & \cellcolor[HTML]{C3E2C3}\textbf{0.7841} & 0.6119 & 0.3384 & 0.4361 & 0.8410 &  & 0.9551 & 0.9304 & 0.8843 & 0.9067 & 0.9891 \\ \hline
\textbf{CatBoost} & \cellcolor[HTML]{C3E2C3}\textbf{0.7841} & 0.6119 & 0.3384 & 0.4361 & 0.8418 &  & 0.9531 & 0.9224 & 0.8843 & 0.9029 & 0.9906 \\ \hline
\textbf{LightGBM} & \cellcolor[HTML]{C3E2C3}\textbf{0.7841} & 0.6119 & 0.3384 & 0.4361 & \cellcolor[HTML]{C3E2C3}\textbf{0.8421} &  & 0.9470 & 0.9203 & 0.8595 & 0.8888 & 0.9903 \\ \hline
\textbf{ExtraTree} & \cellcolor[HTML]{C3E2C3}\textbf{0.7841} & 0.6119 & 0.3384 & 0.4361 & 0.8400 &  & 0.9551 & \cellcolor[HTML]{C3E2C3}\textbf{0.9380} & 0.8760 & 0.9059 & 0.9868 \\ \hline
\end{tabular}
}
\end{table*}

\begin{table*}[!htbp]
\caption{Embedding Aidded Traffic Congestion Classification Results}
\renewcommand{\arraystretch}{1.2}
\label{tab:res2}
\centering
\resizebox{.99\linewidth}{!}{
\begin{tabular}{llccccclccccc}
\hline
\multicolumn{1}{c}{} & \multicolumn{1}{c}{} & \multicolumn{5}{c}{\textbf{wo/ feature enginnering}} &  & \multicolumn{5}{c}{\textbf{w/ feature enginnering}} \\ \cline{3-13} 
\multicolumn{1}{c}{\multirow{-2}{*}{\textbf{\begin{tabular}[c]{@{}c@{}}Embedding \\ Model\end{tabular}}}} & \multicolumn{1}{c}{\multirow{-2}{*}{\textbf{\begin{tabular}[c]{@{}c@{}}Classification\\ Model\end{tabular}}}} & \textbf{Acc.} & \textbf{Precision} & \textbf{Recall} & \textbf{F1} & \textbf{AUROC} &  & \textbf{Acc.} & \textbf{Precision} & \textbf{Recall} & \textbf{F1} & \textbf{AUROC} \\ \hline
 & \textbf{LR} & 0.7657 & 0.5517 & \cellcolor[HTML]{E2FFE2}0.2644 & \cellcolor[HTML]{E2FFE2}0.3575 & \cellcolor[HTML]{E2FFE2}0.8085 &  & \cellcolor[HTML]{E2FFE2}0.9633 & \cellcolor[HTML]{E2FFE2}0.9401 & \cellcolor[HTML]{E2FFE2}0.9090 & \cellcolor[HTML]{E2FFE2}0.9243 & 0.9912 \\ \cline{2-13} 
 & \textbf{KNN} & \cellcolor[HTML]{E2FFE2}0.8329 & \cellcolor[HTML]{E2FFE2}0.6560 & \cellcolor[HTML]{E2FFE2}0.6776 & \cellcolor[HTML]{E2FFE2}0.6666 & \cellcolor[HTML]{E2FFE2}0.8694 &  & 0.8655 & 0.7165 & 0.7520 & 0.7338 & 0.9059 \\ \cline{2-13} 
 & \textbf{RF} & \cellcolor[HTML]{C3E2C3}\textbf{0.8513} & \cellcolor[HTML]{C3E2C3}\textbf{0.7222} & \cellcolor[HTML]{E2FFE2}0.6446 & \cellcolor[HTML]{C3E2C3}\textbf{0.6812} & \cellcolor[HTML]{E2FFE2}0.9145 &  & \cellcolor[HTML]{E2FFE2}0.9592 & \cellcolor[HTML]{E2FFE2}0.9469 & 0.8843 & \cellcolor[HTML]{E2FFE2}0.9145 & \cellcolor[HTML]{E2FFE2}0.9856 \\ \cline{2-13} 
 & \textbf{XGBoost} & \cellcolor[HTML]{E2FFE2}0.8492 & \cellcolor[HTML]{E2FFE2}0.7117 & \cellcolor[HTML]{E2FFE2}0.6528 & \cellcolor[HTML]{E2FFE2}0.6810 & \cellcolor[HTML]{E2FFE2}0.9150 &  & \cellcolor[HTML]{E2FFE2}0.9613 & \cellcolor[HTML]{E2FFE2}0.9322 & \cellcolor[HTML]{E2FFE2}0.9090 & \cellcolor[HTML]{E2FFE2}0.9205 & \cellcolor[HTML]{E2FFE2}0.9901 \\ \cline{2-13} 
 & \textbf{CatBoost} & \cellcolor[HTML]{E2FFE2}0.8492 & \cellcolor[HTML]{E2FFE2}0.7156 & \cellcolor[HTML]{E2FFE2}0.6446 & \cellcolor[HTML]{E2FFE2}0.6782 & \cellcolor[HTML]{C3E2C3}\textbf{0.9183} &  & \cellcolor[HTML]{E2FFE2}0.9613 & \cellcolor[HTML]{E2FFE2}0.9473 & \cellcolor[HTML]{E2FFE2}0.8925 & \cellcolor[HTML]{E2FFE2}0.9191 & 0.9904 \\ \cline{2-13} 
 & \textbf{LightGBM} & \cellcolor[HTML]{E2FFE2}0.8391 & \cellcolor[HTML]{E2FFE2}0.6981 & \cellcolor[HTML]{E2FFE2}0.6115 & \cellcolor[HTML]{E2FFE2}0.6519 & \cellcolor[HTML]{E2FFE2}0.9081 &  & \cellcolor[HTML]{E2FFE2}0.9613 & \cellcolor[HTML]{E2FFE2}0.9250 & \cellcolor[HTML]{C3E2C3}\textbf{0.9173} & \cellcolor[HTML]{E2FFE2}0.9211 & \cellcolor[HTML]{E2FFE2}0.9913 \\ \cline{2-13} 
\multirow{-7}{*}{\textbf{Node2Vec}} & \textbf{ExtraTree} & \cellcolor[HTML]{E2FFE2}0.8411 & \cellcolor[HTML]{E2FFE2}0.6869 & \cellcolor[HTML]{E2FFE2}0.6528 & \cellcolor[HTML]{E2FFE2}0.6694 & \cellcolor[HTML]{E2FFE2}0.9056 &  & \cellcolor[HTML]{E2FFE2}0.9613 & \cellcolor[HTML]{E2FFE2}0.9473 & \cellcolor[HTML]{E2FFE2}0.8925 & \cellcolor[HTML]{E2FFE2}0.9191 & \cellcolor[HTML]{E2FFE2}0.9897 \\ \hline
 & \textbf{LR} & 0.7739 & 0.6250 & 0.2066 & 0.3105 & \cellcolor[HTML]{E2FFE2}0.8016 &  & 0.9572 & 0.9310 & 0.8925 & \cellcolor[HTML]{E2FFE2}0.9113 & \cellcolor[HTML]{C3E2C3}\textbf{0.9922} \\ \cline{2-13} 
 & \textbf{KNN} & \cellcolor[HTML]{E2FFE2}0.8309 & \cellcolor[HTML]{E2FFE2}0.6727 & \cellcolor[HTML]{E2FFE2}0.6115 & \cellcolor[HTML]{E2FFE2}0.6406 & \cellcolor[HTML]{E2FFE2}0.8321 &  & 0.9490 & 0.9000 & 0.8925 & 0.8962 & \cellcolor[HTML]{E2FFE2}0.9650 \\ \cline{2-13} 
 & \textbf{RF} & \cellcolor[HTML]{E2FFE2}0.8268 & \cellcolor[HTML]{E2FFE2}0.6525 & \cellcolor[HTML]{E2FFE2}0.6363 & \cellcolor[HTML]{E2FFE2}0.6443 & \cellcolor[HTML]{E2FFE2}0.8798 &  & 0.9490 & 0.9137 & \cellcolor[HTML]{E2FFE2}0.8760 & 0.8945 & \cellcolor[HTML]{E2FFE2}0.9868 \\ \cline{2-13} 
 & \textbf{XGBoost} & \cellcolor[HTML]{E2FFE2}0.8146 & \cellcolor[HTML]{E2FFE2}0.6293 & \cellcolor[HTML]{E2FFE2}0.6033 & \cellcolor[HTML]{E2FFE2}0.6160 & \cellcolor[HTML]{E2FFE2}0.8798 &  & 0.9429 & 0.9043 & 0.8595 & 0.8813 & 0.9870 \\ \cline{2-13} 
 & \textbf{CatBoost} & \cellcolor[HTML]{E2FFE2}0.8126 & \cellcolor[HTML]{E2FFE2}0.6435 & \cellcolor[HTML]{E2FFE2}0.5371 & \cellcolor[HTML]{E2FFE2}0.5855 & \cellcolor[HTML]{E2FFE2}0.8931 &  & 0.9531 & 0.9224 & 0.8843 & 0.9029 & 0.9901 \\ \cline{2-13} 
 & \textbf{LightGBM} & \cellcolor[HTML]{E2FFE2}0.8289 & \cellcolor[HTML]{E2FFE2}0.6554 & \cellcolor[HTML]{E2FFE2}0.6446 & \cellcolor[HTML]{E2FFE2}0.6500 & \cellcolor[HTML]{E2FFE2}0.8928 &  & \cellcolor[HTML]{E2FFE2}0.9511 & 0.9145 & \cellcolor[HTML]{E2FFE2}0.8843 & \cellcolor[HTML]{E2FFE2}0.8991 & 0.9885 \\ \cline{2-13} 
\multirow{-7}{*}{\textbf{GLEE}} & \textbf{ExtraTree} & \cellcolor[HTML]{E2FFE2}0.8370 & \cellcolor[HTML]{E2FFE2}0.6614 & \cellcolor[HTML]{C3E2C3}\textbf{0.6942} & \cellcolor[HTML]{E2FFE2}0.6774 & \cellcolor[HTML]{E2FFE2}0.8603 &  & \cellcolor[HTML]{C3E2C3}\textbf{0.9653} & \cellcolor[HTML]{C3E2C3}\textbf{0.9482} & \cellcolor[HTML]{E2FFE2}0.9090 & \cellcolor[HTML]{C3E2C3}\textbf{0.9282} & \cellcolor[HTML]{E2FFE2}0.9877 \\ \hline
 & \textbf{LR} & \cellcolor[HTML]{E2FFE2}0.8228 & \cellcolor[HTML]{E2FFE2}0.7073 & \cellcolor[HTML]{E2FFE2}0.4793 & \cellcolor[HTML]{E2FFE2}0.5714 & \cellcolor[HTML]{E2FFE2}0.8405 &  & 0.9572 & \cellcolor[HTML]{E2FFE2}0.9464 & 0.8760 & 0.9098 & 0.9910 \\ \cline{2-13} 
 & \textbf{KNN} & \cellcolor[HTML]{E2FFE2}0.8350 & \cellcolor[HTML]{E2FFE2}0.6666 & \cellcolor[HTML]{E2FFE2}0.6611 & \cellcolor[HTML]{E2FFE2}0.6639 & \cellcolor[HTML]{E2FFE2}0.8670 &  & 0.9307 & 0.8320 & \cellcolor[HTML]{E2FFE2}0.9008 & 0.8650 & 0.9560 \\ \cline{2-13} 
 & \textbf{RF} & \cellcolor[HTML]{E2FFE2}0.8411 & \cellcolor[HTML]{E2FFE2}0.7009 & \cellcolor[HTML]{E2FFE2}0.6198 & \cellcolor[HTML]{E2FFE2}0.6578 & \cellcolor[HTML]{E2FFE2}0.8833 &  & 0.9511 & 0.9145 & 0.8843 & 0.8991 & \cellcolor[HTML]{E2FFE2}0.9860 \\ \cline{2-13} 
 & \textbf{XGBoost} & \cellcolor[HTML]{E2FFE2}0.8431 & \cellcolor[HTML]{E2FFE2}0.6964 & \cellcolor[HTML]{E2FFE2}0.6446 & \cellcolor[HTML]{E2FFE2}0.6695 & \cellcolor[HTML]{E2FFE2}0.8947 &  & 0.9490 & 0.9000 & \cellcolor[HTML]{E2FFE2}0.8925 & 0.8962 & 0.9875 \\ \cline{2-13} 
 & \textbf{CatBoost} & \cellcolor[HTML]{E2FFE2}0.8370 & \cellcolor[HTML]{E2FFE2}0.7113 & \cellcolor[HTML]{E2FFE2}0.5702 & \cellcolor[HTML]{E2FFE2}0.6330 & \cellcolor[HTML]{E2FFE2}0.9075 &  & 0.9490 & 0.9137 & 0.8760 & 0.8945 & 0.9901 \\ \cline{2-13} 
 & \textbf{LightGBM} & \cellcolor[HTML]{E2FFE2}0.8411 & \cellcolor[HTML]{E2FFE2}0.7128 & \cellcolor[HTML]{E2FFE2}0.5950 & \cellcolor[HTML]{E2FFE2}0.6486 & \cellcolor[HTML]{E2FFE2}0.9011 &  & \cellcolor[HTML]{E2FFE2}0.9551 & \cellcolor[HTML]{E2FFE2}0.9230 & \cellcolor[HTML]{E2FFE2}0.8925 & \cellcolor[HTML]{E2FFE2}0.9075 & 0.9895 \\ \cline{2-13} 
\multirow{-7}{*}{\textbf{NETMF}} & \textbf{ExtraTree} & \cellcolor[HTML]{E2FFE2}0.8329 & \cellcolor[HTML]{E2FFE2}0.6695 & \cellcolor[HTML]{E2FFE2}0.6363 & \cellcolor[HTML]{E2FFE2}0.6525 & \cellcolor[HTML]{E2FFE2}0.8726 &  & \cellcolor[HTML]{E2FFE2}0.9592 & 0.9316 & \cellcolor[HTML]{E2FFE2}0.9008 & \cellcolor[HTML]{E2FFE2}0.9159 & \cellcolor[HTML]{E2FFE2}0.9889 \\ \hline
 & \textbf{LR} & 0.7739 & 0.6250 & 0.2066 & 0.3105 & \cellcolor[HTML]{E2FFE2}0.8038 &  & 0.9572 & 0.9310 & 0.8925 & 0.9113 & 0.9922 \\ \cline{2-13} 
 & \textbf{KNN} & \cellcolor[HTML]{E2FFE2}0.7922 & \cellcolor[HTML]{E2FFE2}0.5887 & \cellcolor[HTML]{E2FFE2}0.5206 & \cellcolor[HTML]{E2FFE2}0.5526 & \cellcolor[HTML]{E2FFE2}0.8005 &  & 0.9490 & 0.9000 & 0.8925 & 0.8962 & 0.9625 \\ \cline{2-13} 
 & \textbf{RF} & \cellcolor[HTML]{E2FFE2}0.7881 & 0.5794 & \cellcolor[HTML]{E2FFE2}0.5124 & \cellcolor[HTML]{E2FFE2}0.5438 & \cellcolor[HTML]{E2FFE2}0.8563 &  & \cellcolor[HTML]{E2FFE2}0.9592 & \cellcolor[HTML]{E2FFE2}0.9391 & 0.8925 & \cellcolor[HTML]{E2FFE2}0.9152 & \cellcolor[HTML]{E2FFE2}0.9863 \\ \cline{2-13} 
 & \textbf{XGBoost} & \cellcolor[HTML]{E2FFE2}0.8044 & \cellcolor[HTML]{E2FFE2}0.6213 & \cellcolor[HTML]{E2FFE2}0.5289 & \cellcolor[HTML]{E2FFE2}0.5714 & \cellcolor[HTML]{E2FFE2}0.8557 &  & 0.9531 & 0.9224 & 0.8843 & 0.9029 & 0.9885 \\ \cline{2-13} 
 & \textbf{CatBoost} & \cellcolor[HTML]{E2FFE2}0.7983 & 0.6100 & \cellcolor[HTML]{E2FFE2}0.5041 & \cellcolor[HTML]{E2FFE2}0.5520 & \cellcolor[HTML]{E2FFE2}0.8707 &  & \cellcolor[HTML]{E2FFE2}0.9613 & \cellcolor[HTML]{E2FFE2}0.9473 & \cellcolor[HTML]{E2FFE2}0.8925 & \cellcolor[HTML]{E2FFE2}0.9191 & 0.9900 \\ \cline{2-13} 
 & \textbf{LightGBM} & \cellcolor[HTML]{E2FFE2}0.8309 & \cellcolor[HTML]{E2FFE2}0.6862 & \cellcolor[HTML]{E2FFE2}0.5785 & \cellcolor[HTML]{E2FFE2}0.6278 & \cellcolor[HTML]{E2FFE2}0.8836 &  & \cellcolor[HTML]{E2FFE2}0.9531 & \cellcolor[HTML]{E2FFE2}0.9298 & \cellcolor[HTML]{E2FFE2}0.8760 & \cellcolor[HTML]{E2FFE2}0.9021 & 0.9898 \\ \cline{2-13} 
\multirow{-7}{*}{\textbf{GraphWave}} & \textbf{ExtraTree} & \cellcolor[HTML]{E2FFE2}0.7881 & 0.5794 & \cellcolor[HTML]{E2FFE2}0.5124 & \cellcolor[HTML]{E2FFE2}0.5438 & \cellcolor[HTML]{E2FFE2}0.8530 &  & \cellcolor[HTML]{E2FFE2}0.9633 & \cellcolor[HTML]{E2FFE2}0.9478 & \cellcolor[HTML]{E2FFE2}0.9008 & \cellcolor[HTML]{E2FFE2}0.9237 & 0.9834 \\ \hline
\end{tabular}
}
\end{table*}

\subsection{Road Traffic Prediction: ExtraTrees}
ExtraTrees is an ensemble model based on decision trees. ExtraTrees differ from standard bagging models in terms of tunning of \textit{best splitting} hyperparameter. ExtraTrees chooses the optimum split randomly which makes it much faster.

\section{Experimental Results}\label{sec:5} 

We conducted our experiments using four node embedding methods, three proximity-preserving (Node2Vec, GLEE, NetMF), one structural node-level embedding (GraphWave), and seven classifiers. Next, we give details of these models used in the experiments and road traffic prediction results obtained.

\textbf{Node2Vec} \cite{grover2016node2vec} algorithm learns representations of nodes (i.e. sensors) via random walk procedure. The algorithm creates node embeddings of the graph following a flexible neighborhood sampling strategy that trades off between Breadth-First Search (BFS) and Depth-First Search (DFS). To illustrate, starting from a source node $u$, Node2Vec simulates a random walk of length $l$. Initially, $c_0=u$ represents the starting node, and $c_i$ denotes the $i$ th node in the walk. The nodes $c_i$ are generated based on the probability distribution in Equation \ref{eq_n2v_1}. In this equation, $\pi_{x v}$ is the unnormalized transition probability between nodes $v$ and $x$, and $Z$ is the normalizing constant.

\begin{equation}
P\left(c_i=x \mid c_{i-1}=v\right)=\left\{\begin{array}{cc}
\frac{\pi_{v x}}{Z} & \text { if }(v, x) \in E \\
0 & \text { otherwise }
\end{array}\right\}
\label{eq_n2v_1}
\end{equation}

A $2^{n d}$ order random walk on a graph is defined by two parameters $p$ (return parameter) and $q$ (in-out parameter) which guide the walker in deciding the next step. Consider a scenario in which the walker has just taken a step from node $t$ to node $v$. At this point, the walker evaluates the transition probabilities $\pi_{vx}$ on edges $(v,x)$ leading from $v$. The transition probability here is unnormalized and can be represented as $\pi_{v x}=\alpha_{p q} \cdot w_{v x}$ where the $\alpha_{p q}$, which defined in Equation \ref{eq_n2v_2}, is the length of the shortest path between nodes $t$ and $x$.

\begin{equation}
\alpha_{p q}(t, x)= \begin{cases}\frac{1}{p} & \text { if } d_{t x}=0 \\ 1 & \text { if } d_{t x}=1 \\ \frac{1}{q} & \text { if } d_{t x}=2\end{cases}
\label{eq_n2v_2}
\end{equation}

The parameters $p$ and $q$ determine how the random walk behaves, like either BFS or DFS. Parameter $p$ influences the likelihood of revisiting a node immediately after leaving it. Higher values of $p$ reduce this likelihood, which means higher values for $p$ encourage the walker to make more exploration. The other parameter $q$ determines whether the walker searches for inward or outward nodes. The two parameters are intertwined, meaning the value of one does not alone tell about the behavior of the random walk.

%\cite{Qiu_2018}
\textbf{NetMF} \cite{Qiu_2018} unifies various network embedding models, namely DeepWalk, LINE, PTE, and Node2Vec into an implicit matrix factorization framework, showing their theoretical connections to graph Laplacians. DeepWalk generates random walks on the network similar to how sentences are formed in a language. The NetMF aims to maximize the probability of observing these node sequences, which can be represented as matrix factorization of the transition probability matrix \(\textbf{M}\) in Equation \ref{eq:deepwalk}, where \(U\) and \(V\) are the node embeddings. LINE (Large-scale Information Network Embedding) focuses on preserving both first-order and second-order proximities. First-order proximity ensures that neighbor nodes or nodes with strong connections are closer to each other in the embedding space, while second-order proximity can be interpreted as nodes sharing the same neighbors are more likely to be similar.

\begin{equation}
\label{eq:deepwalk}
\min_{U, V} \| M - UV^\top \|^2,
\end{equation}

\begin{equation}
\label{eq:pte}
\min_{U, V} \sum_{t} \sum_{(i,j) \in E_t} w_{ij}^t \| u_i^t - v_j^t \|^2,
\end{equation}

\begin{equation}
\label{eq:node2vec}
\min_{U, V} \| M_{\text{node2vec}} - UV^\top \|^2.
\end{equation}

\begin{equation}
\label{eq:unified}
\min_{U, V} \| M - UV^\top \|^2,
\end{equation}

PTE (Predictive Text Embedding) is an extension of LINE in heterogeneous networks, optimizing a similar objective for each type of edge and combining them as in Equation \ref{eq:pte} where \(t\) indexes different types of edges. Node2vec extends DeepWalk by introducing a biased random walk to capture homophily and structural equivalence. All these methods can be unified under the general matrix factorization framework in Equation \ref{eq:unified} where \(M\) represents different transition probability matrices or proximity matrices depending on the method used. 

\textbf{GraphWave} \cite{Donnat_2018} is used to learn the neighborhood of each node in a network using a low-dimensional embedding by leveraging heat wavelet diffusion patterns. Let $\textbf{U}$ be the eigenvector matrix from the decomposition of the unnormalized graph Laplacian $\textbf{L}=\textbf{D}-\textbf{A}$, where $\textbf{L}=\textbf{U} \Lambda \textbf{U}^T$, $\Lambda$ is a diagonal matrix of eigenvalues $\lambda$ and $\lambda_1<\lambda_2 \leq \ldots \leq \lambda_N$ are the eigenvalues of $\textbf{L}$. Here, $\textbf{D}$ is the diagonal matrix whose entries are the degrees of each node, and $\textbf{A} = (a_{ij})$ is the edge weight matrix such that $a_{ij}$ represents the weight of the edge between nodes $i$ and $j$. Let $g_s$ be a filter kernel with scaling parameter $s$. The spectral graph wavelet of $g_s$ is the signal resulting from the modulation in the spectral domain of a Dirac signal centered around node $a$. The spectral graph wavelet $\Psi_a$ is given as an $N$-dimensional vector as in Equation \ref{eq_gw_0}, where $\delta_a=\mathbb{1}(a)$ is the one-hot vector of node $a$. Spectral graph wavelets are based on the relationship between the temporal frequencies of a signal and the Laplacian's eigenvalues.

\begin{equation}
\Psi_a=\textbf{U} \operatorname{Diag}\left(g_s\left(\lambda_1\right), \ldots, g_s\left(\lambda_N\right)\right) \textbf{U}^T \delta_a
\label{eq_gw_0}
\end{equation}

\begin{equation}
\phi_X(t)=\mathbb{E}\left[e^{i t X}\right]
\label{eq_gw_1}
\end{equation}

\begin{equation}
\Psi_a(t)=\frac{1}{N} \sum_{m=1}^N e^{i t \Psi_{m a}}
\label{eq_gw_2}
\end{equation}

\begin{equation}
\chi_a=\left[\operatorname{Re}\left(\phi_a\left(t_i\right)\right), \operatorname{Im}\left(\phi_a\left(t_i\right)\right)\right]
\label{eq_gw_3}
\end{equation}

For every node $a$, GraphWave computes a $2 d$-dimensional vector $\chi_a$. To obtain diffusion patterns of every node, spectral graph wavelets are used. Diffusion patterns are stored in a ma$\operatorname{trix} \Psi . \Psi$ is a $N x N$ matrix, where the $a$-th column is the spectral graph wavelet for a heat kernel centered at node $a$. GraphWave embeds spectral graph wavelet coefficient distributions into $2 d$ dimensional space by calculating the characteristic function for each node's coefficients $\Psi_a$ and samples it at $d$ evenly spaced points. The characteristic function of a probability distribution $X$ is as in Equation \ref{eq_gw_1}. For a given node $a$ and scale $s$, the empirical characteristic function of $\Psi_a$ is defined as in Equation \ref{eq_gw_2}. Embedding $\chi_a$ of node $a$ is obtained by sampling the parametric function in Equation \ref{eq_gw_1} at $d$ evenly spaced points $t_1 \ldots t_d$ and concatenating the values as in Equation \ref{eq_gw_3}.

\textbf{Logistic Regression (LR)} uses the sigmoid function $\sigma(x) = 1/{(1 + e^{-x})}$ to transform the weighted sum of input features into a probability value ranging between zero and one.

\textbf{K-Nearest Neighbor (KNN)} classifies data points based on their proximity to other data points.

\textbf{Random Forest (RF)} is a bootstrap ensemble model. It creates several decision trees on data samples and then selects the best solution using voting. 

\textbf{XGBoost} is an implementation of gradient boosting. The key idea behind the XGBoost is the improvement of speed and performance using the reasons behind the good performance such as regularization and handling sparse data.

\textbf{LightGBM} is a histogram-based gradient boosting algorithm reducing the computational cost by converting continuous variables into discrete ones. Since the training time of the decision trees is directly proportional to the number of computations, hence the number of splits, LightGBM provides a shorter model training time and efficient resource utilization.

\textbf{Catboost} is a gradient-boosting ML algorithm known for its high predictive accuracy and speed. It employs techniques such as ordered boosting and oblivious trees to handle various data types effectively and mitigate overfitting. 

\subsection{Traffic Congestion Prediction Results}

In this study, we approached the traffic congestion prediction problem as a binary classification task, where the output for each sensor node determines whether congestion may occur or not. A straightforward and commonly used method for labeling involves directly using average speed values, as shown in Equation \ref{eq:density1}. However, given the narrow streets in urban areas where the average speed is low but congestion is absent, we conducted our experiments using an alternative formulation. This formulation also takes into account the number of passing vehicles, as represented in Equation \ref{eq:density2}. The function \(C\) gives the proper label on congestion for each node in a timestamp where the \(s\) is the speed and \(v\) is the number of vehicles. Additionally, \(\sigma\) is the sigmoid function mapping the output between 0 and 1 and the \(\tau\) is the threshold value ranging between 0 and 1.

\begin{equation}
C(s) = 
\begin{cases} 
0, & \text{if } \frac{1}{e^s}  < \tau \\
1, & \text{otherwise}
\end{cases}
\label{eq:density1}
\end{equation}

\begin{equation}
C(s,v) = 
\begin{cases} 
0, & \text{if } \sigma\left(\frac{s}{v}\right) < \tau \\
1, & \text{otherwise }
\end{cases}
\label{eq:density2}
\end{equation}

The results obtained with various classifier models without adding node embedding vectors to the data are presented in Table \ref{tab:res1}, while the results with node embedding are presented in Table \ref{tab:res2}. In both Table \ref{tab:res1} and Table \ref{tab:res2}, the value of feature engineering studies was also queried. The values highlighted in Table \ref{tab:res1} indicate the most accurate value for the relevant metric. 

Comparing the 'without feature engineering' and 'with feature engineering' cases, it is seen that the traffic density one hour ago and the traffic density one week ago features obtained via feature engineering studies are quite effective. On the other hand, in Table \ref{tab:res2}, the light green color indicates that node embeddings provide a performance improvement compared to the result in Table \ref{tab:res1} for that metric, classifier and w/wo feature cases, while dark greens indicate the highest scores in that category. When the results in Table \ref{tab:res1} and Table \ref{tab:res2} are analyzed together, for the scenario 'without feature engineering', the approach of adding node embedding vectors to the data improves the prediction performance regardless of the classifier and node embedding methods. Considering all the findings together, it is seen that the prominent approach (with Acc.= 0.9653, Precision= 0.9482, Recall= 0.9090, F1= 0.9282, AUROC= 0.9877) is the one that strengthens temporal links through feature engineering, node embedding with GLEE to represent inter-related relationships within the traffic network and road traffic congestion prediction with ExtraTrees.

\section{Conclusion and Future Works} \label{sec:6} 

In this study, we propose a novel Road Traffic Prediction Model that effectively leverages traffic flow features—such as average speed, vehicle count, min. speed, and max. speed—and sensor features like geographical categorizations (continents, districts, highways/urban areas, and population density), which are visualized through maps. The inclusion of temporal data w/feature engineering studies, such as traffic densities from the previous week and hour, further enhances prediction accuracy. Node embeddings obtained via the GLEE, combined with the ExtraTrees classifier, enable the distinction between normal and congested traffic more accurately. Analysis of results shows significant performance improvements when using feature engineering and node embeddings, highlighting the importance of these techniques in traffic prediction. For future work, we plan to work on traffic prediction for dynamic graphs.

\section*{Acknowledgements}
This research is supported by the Scientific and Technological Research Council of Turkey (TUBITAK) 1515 Frontier R\&D Laboratories Support Program (project number 5239903) and the ITU Scientific Research Projects Fund under grant number YESAP-2024-45920.

\bibliographystyle{IEEEtran_bib}
\bibliography{IEEEabrv, bibliography}

\begin{thebibliography}{10}
\providecommand{\url}[1]{#1}
\csname url@samestyle\endcsname
\providecommand{\newblock}{\relax}
\providecommand{\bibinfo}[2]{#2}
\providecommand{\BIBentrySTDinterwordspacing}{\spaceskip=0pt\relax}
\providecommand{\BIBentryALTinterwordstretchfactor}{4}
\providecommand{\BIBentryALTinterwordspacing}{\spaceskip=\fontdimen2\font plus
\BIBentryALTinterwordstretchfactor\fontdimen3\font minus \fontdimen4\font\relax}
\providecommand{\BIBforeignlanguage}[2]{{%
\expandafter\ifx\csname l@#1\endcsname\relax
\typeout{** WARNING: IEEEtran.bst: No hyphenation pattern has been}%
\typeout{** loaded for the language `#1'. Using the pattern for}%
\typeout{** the default language instead.}%
\else
\language=\csname l@#1\endcsname
\fi
#2}}
\providecommand{\BIBdecl}{\relax}
\BIBdecl

\bibitem{10077454}
S.~Rahmani, A.~Baghbani, N.~Bouguila, and Z.~Patterson, ``Graph neural networks for intelligent transportation systems: A survey,'' \emph{IEEE Transactions on Intelligent Transportation Systems}, vol.~24, no.~8, pp. 8846--8885, 2023.

\bibitem{9046288}
Z.~Wu, S.~Pan, F.~Chen, G.~Long, C.~Zhang, and P.~S. Yu, ``A comprehensive survey on graph neural networks,'' \emph{IEEE Transactions on Neural Networks and Learning Systems}, vol.~32, no.~1, pp. 4--24, 2021.

\bibitem{jin2023survey}
M.~Jin, H.~Y. Koh, Q.~Wen, D.~Zambon, C.~Alippi, G.~I. Webb, I.~King, and S.~Pan, ``A survey on graph neural networks for time series: Forecasting, classification, imputation, and anomaly detection,'' 2023.

\bibitem{yin}
X.~Yin, G.~Wu, J.~Wei, Y.~Shen, H.~Qi, and B.~Yin, ``Deep learning on traffic prediction: Methods, analysis, and future directions,'' \emph{IEEE Transactions on Intelligent Transportation Systems}, vol.~23, no.~6, pp. 4927--4943, 2021.

\bibitem{Torres_2020}
\BIBentryALTinterwordspacing
L.~Torres, K.~S. Chan, and T.~Eliassi-Rad, ``Glee: Geometric laplacian eigenmap embedding,'' \emph{Journal of Complex Networks}, vol.~8, no.~2, Mar. 2020. [Online]. Available: \url{http://dx.doi.org/10.1093/comnet/cnaa007}
\BIBentrySTDinterwordspacing

\bibitem{narayanan2017graph2vec}
A.~Narayanan, M.~Chandramohan, R.~Venkatesan, L.~Chen, Y.~Liu, and S.~Jaiswal, ``graph2vec: Learning distributed representations of graphs,'' 2017.

\bibitem{Perozzi_2014}
\BIBentryALTinterwordspacing
B.~Perozzi, R.~Al-Rfou, and S.~Skiena, ``Deepwalk: online learning of social representations,'' in \emph{Proceedings of the 20th ACM SIGKDD international conference on Knowledge discovery and data mining}, ser. KDD ’14.\hskip 1em plus 0.5em minus 0.4em\relax ACM, Aug. 2014. [Online]. Available: \url{http://dx.doi.org/10.1145/2623330.2623732}
\BIBentrySTDinterwordspacing

\bibitem{Donnat_2018}
\BIBentryALTinterwordspacing
C.~Donnat, M.~Zitnik, D.~Hallac, and J.~Leskovec, ``Learning structural node embeddings via diffusion wavelets,'' in \emph{Proceedings of the 24th ACM SIGKDD International Conference on Knowledge Discovery and Data Mining}, ser. KDD ’18.\hskip 1em plus 0.5em minus 0.4em\relax ACM, Jul. 2018. [Online]. Available: \url{http://dx.doi.org/10.1145/3219819.3220025}
\BIBentrySTDinterwordspacing

\bibitem{li2018diffusionconvolutionalrecurrentneural}
\BIBentryALTinterwordspacing
Y.~Li, R.~Yu, C.~Shahabi, and Y.~Liu, ``Diffusion convolutional recurrent neural network: Data-driven traffic forecasting,'' 2018. [Online]. Available: \url{https://arxiv.org/abs/1707.01926}
\BIBentrySTDinterwordspacing

\bibitem{liu2023largest}
X.~Liu, Y.~Xia, Y.~Liang, J.~Hu, Y.~Wang, L.~Bai, C.~Huang, Z.~Liu, B.~Hooi, and R.~Zimmermann, ``Largest: A benchmark dataset for large-scale traffic forecasting,'' 2023.

\bibitem{sun2021spatio}
B.~Sun, T.~Sun, and P.~Jiao, ``Spatio-temporal segmented traffic flow prediction with anprs data based on improved xgboost,'' \emph{Journal of Advanced Transportation}, vol. 2021, no.~1, p. 5559562, 2021.

\bibitem{bai2021prepct}
M.~Bai, Y.~Lin, M.~Ma, P.~Wang, and L.~Duan, ``Prepct: Traffic congestion prediction in smart cities with relative position congestion tensor,'' \emph{Neurocomputing}, vol. 444, pp. 147--157, 2021.

\bibitem{tran2022predicting}
M.-D. Tran and N.~M. Saleem, ``Predicting high-risk congestion areas during heavy rain using multi prediction model and maximum periodic frequent pattern algorithms,'' in \emph{Proceedings of the 3rd ACM Workshop on Intelligent Cross-Data Analysis and Retrieval}, 2022, pp. 27--31.

\bibitem{wang2021freeway}
F.~Wang, H.~Cheng, H.~Dai, and H.~Han, ``Freeway short-term travel time prediction based on lightgbm algorithm,'' in \emph{IOP Conference Series: Earth and Environmental Science}, vol. 638, no.~1.\hskip 1em plus 0.5em minus 0.4em\relax IOP Publishing, 2021, p. 012029.

\bibitem{Song_Lin_Guo_Wan_2020}
C.~Song, Y.~Lin, S.~Guo, and H.~Wan, ``Spatial-temporal synchronous graph convolutional networks: A new framework for spatial-temporal network data forecasting,'' \emph{Proceedings of the AAAI Conference on Artificial Intelligence}, vol.~34, no.~01, pp. 914--921, Apr. 2020.

\bibitem{tuik}
``{TUIK},'' {\sl https://data.tuik.gov.tr/Bulten/Index?p=Motorlu-Kara-Tasitlari-Aralik-2023-49432}.

\bibitem{grover2016node2vec}
A.~Grover and J.~Leskovec, ``node2vec: Scalable feature learning for networks,'' 2016.

\bibitem{Qiu_2018}
\BIBentryALTinterwordspacing
J.~Qiu, Y.~Dong, H.~Ma, J.~Li, K.~Wang, and J.~Tang, ``Network embedding as matrix factorization: Unifying deepwalk, line, pte, and node2vec,'' in \emph{Proceedings of the Eleventh ACM International Conference on Web Search and Data Mining}, ser. WSDM 2018.\hskip 1em plus 0.5em minus 0.4em\relax ACM, Feb. 2018. [Online]. Available: \url{http://dx.doi.org/10.1145/3159652.3159706}
\BIBentrySTDinterwordspacing

\end{thebibliography}
\end{document}